\documentclass[runningheads]{llncs}

\usepackage{eccv}

\usepackage{eccvabbrv}

\usepackage{graphicx}
\usepackage{booktabs}
\usepackage{xcolor, colortbl}

\usepackage[accsupp]{axessibility}  

\usepackage{hyperref}

\usepackage{orcidlink}

\definecolor{LightBlue}{rgb}{0.88,1,1}
\definecolor{babypink}{rgb}{0.96, 0.76, 0.76}
\definecolor{classicrose}{rgb}{0.98, 0.8, 0.91}
\definecolor{piggypink}{rgb}{0.99, 0.87, 0.9}

\begin{document}

\title{Contrastive Learning with Synthetic Positives} 


\author{
Dewen Zeng\inst{1} \and
Yawen Wu\inst{1} \and
Xinrong Hu\inst{1} \and
Xiaowei Xu\inst{2} \and
Yiyu Shi\inst{1}
}

\authorrunning{D Zeng, et al.}

\institute{University of Notre Dame, Notre Dame, IN, USA \\\email{\{dzeng2, ywu37, xhu7, yshi4\}@nd.edu}\and
Guangdong Provincial People's Hospital, Guangzhou, China}

\maketitle

\begin{abstract}
  Contrastive learning with the nearest neighbor has proved to be one of the most efficient self-supervised learning (SSL) techniques by utilizing the similarity of multiple instances within the same class. However, its efficacy is constrained as the nearest neighbor algorithm primarily identifies ``easy'' positive pairs, where the representations are already closely located in the embedding space. In this paper, we introduce a novel approach called Contrastive Learning with Synthetic Positives (CLSP) that utilizes synthetic images, generated by an unconditional diffusion model, as the additional positives to help the model learn from diverse positives. Through feature interpolation in the diffusion model sampling process, we generate images with distinct backgrounds yet similar semantic content to the anchor image. These images are considered ``hard'' positives for the anchor image, and when included as supplementary positives in the contrastive loss, they contribute to a performance improvement of over 2\% and 1\% in linear evaluation compared to the previous NNCLR and All4One methods across multiple benchmark datasets such as CIFAR10, achieving state-of-the-art methods. On transfer learning benchmarks, CLSP outperforms existing SSL frameworks on 6 out of 8 downstream datasets. We believe CLSP establishes a valuable baseline for future SSL studies incorporating synthetic data in the training process. The source code is available at \href{https://github.com/dewenzeng/clsp}{https://github.com/dewenzeng/clsp}.
  \keywords{Self-supervised learning \and Contrastive learning \and Diffusion model}
\end{abstract}

\section{Introduction}
\label{sec:intro}
Over the past few years, contrastive learning \cite{chen2020simple,chen2020big,he2020momentum,chen2021exploring} has become one of the most successful self-supervised learning (SSL) approaches for visual representation learning from unlabeled data. Contrastive learning operates on the principle of instance discrimination, wherein models are encouraged to be invariant to different views of the same sample (a.k.a, positive pair) and distinguishable from views of different images. Because the training scheme does not rely on human annotation, contrastive learning can learn generic representations from extensive unlabeled data. Such representations prove particularly valuable in scenarios where acquiring labels is costly, scarce, or noisy \cite{balestriero2023cookbook}.

The effectiveness of contrastive learning relies on the careful selection or generation of positive and negative pairs. Recent studies have delved into advanced strategies for selecting positive/negative pairs to enhance the efficiency of contrastive learning. In the case of negative pairs, conventional contrastive methods uniformly sample negatives from the training set without considering the informativeness between different samples, potentially hindering the learning process. Utilizing importance scores based on feature similarity \cite{robinson2020contrastive, huynh2022boosting} or uncertainty \cite{ma2020active, tabassum2022hard} are two approaches to selecting more informative ``hard'' negatives. In addition, employing mixup from either the image or representation space \cite{lee2020mix, kim2020mixco, shen2022mix, kalantidis2020hard, zhu2021improving} proves to be another effective means of generating diverse and informative negatives for training. Concerning positive pairs, a key limitation of standard contrastive learning lies in its heavy reliance on the data augmentation pipeline, which may fail to cover all variances in a given class. Strategies like nearest neighbor mining \cite{dwibedi2021little, koohpayegani2021mean, Estepa_2023_ICCV} and generative model \cite{wu2023synthetic} have been proposed to enhance positive diversity beyond basic data augmentations. NNCLR \cite{dwibedi2021little} and MSF \cite{koohpayegani2021mean} propose to use the first and $k$ nearest neighbors as the multiple instances positives. All4One \cite{Estepa_2023_ICCV} improves them by incorporating a centroid contrastive objective to learn contextual information from multiple neighbors. However, the nearest-neighbor strategy can only identify ``easy'' positives, as the model is already adept at mapping them to close representations. It fails to address situations where samples are mapped to the wrong cluster. Using generative adversarial networks (GAN) to generate synthetic positives not present in the original dataset has proved to be another efficient way to increase data diversity \cite{wu2023synthetic}. However, this framework requires simultaneous training of the contrastive learning model and GAN, making it unstable and hard to control the quality of the generated positives.

This paper proposes a new approach called Contrastive Learning with Synthetic Positives (CLSP),  which integrates additional synthetic positives into a conventional contrastive training schema. These synthetic positives exhibit rich deformations and diverse backgrounds, allowing the model to emphasize generic features beyond naive data augmentation. Specifically, our method employs the unconditional diffusion model to generate positives. It has been proved that well-trained unconditional diffusion models can capture semantic information at their intermediate layers \cite{baranchuk2021label, xiang2023denoising}. During a random reversed diffusion process, we can replace the features of the intermediate layers with the semantic features extracted from an anchor image. This results in the generation of images possessing similar semantic content to the anchor image but differing in background and context due to the randomness of features in other layers. We consider these synthetic samples as ``hard'' positives of the anchor image because the semantic similarity could be subtle and extremely distorted. To learn from such positive pairs, we utilize an additional loss term, similar to \cite{grill2020bootstrap}, to minimize their feature similarity. Given the inherent slowness of the diffusion sampling process, generating synthetic positives on the fly is impractical, even with speedup solutions like DDIM \cite{song2020denoising}. Consequently, we pre-generate a positive candidate set with a size of $k$ ($k\leq8$) for each sample in the training set. During training, we randomly sample one image from this set as an additional positive. Experimental results show that CLSP can significantly improve the representation learning performance when compared to its conventional counterpart.

Our contributions can be summarized as follows:
(1) We delve into the sampling process of a pre-trained unconditional diffusion model, revealing that positive samples can be effectively generated by replacing the intermediate features in a random sampling process with the features extracted from the anchor image.
(2) We introduce CLSP, a method designed to integrate the generated positives into the contrastive learning framework, thereby helping the model learn generic presentations. We demonstrate that CLSP increases the linear evaluation performance of conventional contrastive learning methods (e.g., SimCLR, MoCo) on several benchmark datasets (e.g., CIFAR10, CIFAR100) and achieves state-of-the-art (SOTA) performance.
(3) Our method surpasses previous SSL methods on 6 out of 8 transfer learning tasks. Extensive ablation studies affirm the robustness of CLSP across varying model sizes and training iterations.

\section{Related Work}

\textbf{Contrastive Learning.}
Contrastive learning, an effective SSL approach, has shown remarkable success in various domains such as computer vision \cite{ma2020active, pan2021videomoco, zeng2021positional, qian2021spatiotemporal}, natural language processing \cite{aberdam2021sequence, ye2021efficient, wang2022contrastive}, and graph learning \cite{you2020graph, hassani2020contrastive}, rivaling supervised learning models. Various frameworks have been proposed to learn visual representations using contrastive learning. CPC \cite{oord2018representation} predicts the future output of sequential data by using a probabilistic contrastive loss to capture maximally useful information. SimCLR \cite{chen2020simple, chen2020big} uses other samples from the mini-batch as negative pairs. MoCo \cite{he2020momentum, chen2020improved} uses a momentum-updated memory bank to maintain numerous negative samples. Given the necessity of learning from both positive and negative pairs, researchers have proposed several techniques to construct more informative positive and negative pairs to increase training efficiency. For example, \cite{kalantidis2020hard, lee2020mix, kim2020mixco, shen2022mix} apply Mixup \cite{zhang2018mixup} to generate hard negatives. \cite{robinson2020contrastive} proposed a negative sampling strategy to reweight the contrastive learning objective. UnReMix \cite{tabassum2022hard} combines anchor similarity, model uncertainty, and representativeness to sample hard negatives, which improves the results on several visual, text, and graph benchmark datasets. Other works use nearest neighbor to select multi-instance positives to increase the feature diversity within the same semantic class \cite{dwibedi2021little, koohpayegani2021mean, Estepa_2023_ICCV}. However, the nearest neighbor algorithm can only identify easy positives, which makes the improvement marginal when the discrimination task becomes harder. \cite{wu2023synthetic} address this challenge by jointly learning the contrastive learning model and a GAN, a pair of momentum-updated generators are proposed to generate hard positive pairs. However, the joint learning schema is highly unstable and hard to optimize. Other enhancements in contrastive learning involve optimizing the contrastive loss \cite{chuang2020debiased, xie2021propagate, yeh2022decoupled} or eliminating negative pairs such as SwaV \cite{caron2020unsupervised}, SimSiam \cite{chen2021exploring}, BYOL \cite{grill2020bootstrap}, and BarlowTwins \cite{zbontar2021barlow}. These methods don't face false negative issues, as they solely require positive pairs. Our work in this paper is built upon standard contrastive learning (e.g., SimCLR and MoCo) and focuses on the hard positive generation.

\noindent\textbf{Diffusion Model.} Recently, diffusion models \cite{ho2020denoising, karras2022elucidating, rombach2022high, ramesh2022hierarchical, dhariwal2021diffusion} have become the state-of-the-art generative models, demonstrating superiority on various benchmark datasets \cite{ho2020denoising, dhariwal2021diffusion, nichol2021improved, ho2021classifier}. Unlike GANs, diffusion models are stable to train and less prone to collapse. Recent works like DALLE2 \cite{ramesh2022hierarchical} and Stable Diffusion \cite{rombach2022high} have shown exceptional performance in text-to-image generation, unconditional image generation, and super-resolution. \cite{azizi2023synthetic} proved that synthetic data from diffusion models can further increase the accuracy of the supervised models on ImageNet.

Apart from their high-fidelity image generation ability, diffusion models have also shown excellent visual representation learning ability, even in the absence of labeled information. \cite{baranchuk2021label} demonstrates that using the diffusion model as the feature extractor for semantic segmentation significantly boosts the segmentation performance where only a few training samples are provided. \cite{xiang2023denoising} proves that a pre-trained unconditional diffusion model has already learned strongly linear-separable representations at its intermediate layers. RepFusion \cite{yang2023diffusion} extracts features from a pre-trained diffusion model to guide the training of student networks for downstream tasks. Our hard positive generation process heavily relies on the representation learning performance of the unconditional diffusion model. Because the intermediate features are semantically linear-separable in diffusion models, we can use feature interpolation to combine semantic features from random sampling with those extracted from the anchor image to generate positive samples of the anchor image. The positive similarity can be controlled by adjusting the interpolation weight between these two features.

\section{Method}

We begin by introducing the idea of SimCLR because it is conceptually simple. Given a mini-batch of $N$ images $\{x_1, x_2, ..., x_N\}$, for each image $x_i$, we obtain two random augmented views $x_i^1=aug(x_i)$ and $x_i^2=aug(x_i)$, where $aug(\cdot)$ is the random augmentation function. Both views are fed into the same encoder $f(\cdot)$ to generate the embeddings $h_i^1 = f(x_i^1)$ and $h_i^2 = f(x_i^2)$, and then projected to the space where contrastive loss is applied after normalization, resulting in $z_i^1 = ||g(h_i^1)||_2$, $z_i^2 = ||g(h_i^2)||_2$, where $g(\cdot)$ denotes the MLP projector. SimCLR solves the instance discrimination problem by treating augmented views from the same image $x_i^k, k\in\{1,2\}$ as the positive pair and all the other $2N-2$ views in the mini-batch as the negatives. The contrastive loss of view $x_i^1$ is defined as:
\begin{equation}
    \mathcal{L}_{i,k} = - \log \frac{\exp{(z_i^1 \cdot z_i^2 / \tau)}}{\sum_{j\in [1,N], k \in \{1,2\}, i\neq j}{\exp (z_i^1 \cdot z_j^k / \tau)}}.
\end{equation}

\begin{figure}[t!]
	\centering
	\includegraphics[width=1.0\linewidth]{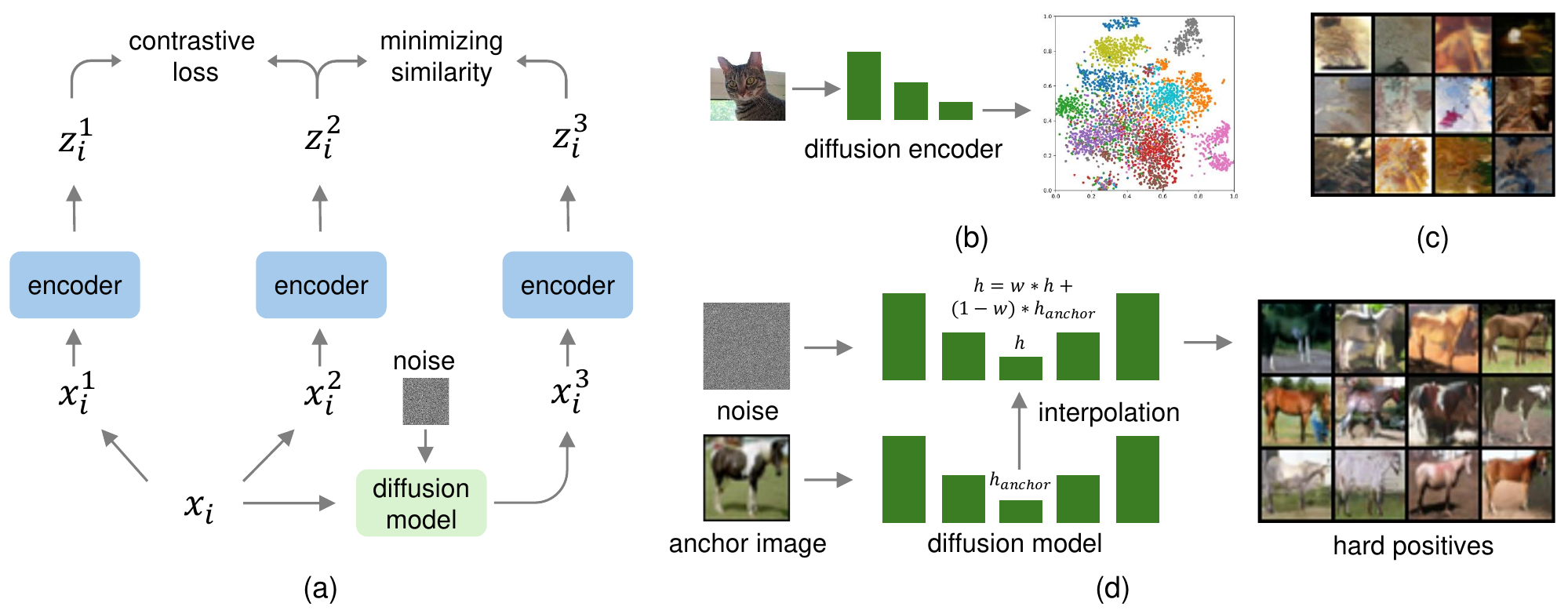}
	\vspace{-10pt}
	\caption{(a) Overview of the proposed CLSP framework, we use a diffusion model to generate an additional positive $x_i^3$ to increase the positive diversity for better representation learning. (b) The t-SNE plot of features extracted from the intermediate layer of the diffusion model trained on CIFAR10. The features are generated at timestamp 50. (c) The generated images only contain background information if intermediate features are masked, suggesting the decoupling of semantic and background information in different layers of the diffusion model. (d) Using feature interpolation to generate hard positives, the generated images contain similar semantic content to the anchor image but differ in context and background.}
	\label{fig:overview}
	\vspace{-10pt}
\end{figure}

where $\tau$ denotes a temperature parameter. The overall loss for the given mini-batch is $\mathcal{L} = \sum_{i \in [1,N], k \in \{1,2\}}\mathcal{L}_{i,k}$.

Since the positive pairs in SimCLR only consider naive data augmentations such as random crops and color changes. It fails to cover the situations where the semantic instance undergoes severe deformation or appears in a different background, which may hinder its generalization ability toward large intra-class variations. To address this problem, we propose to enhance the positive diversity by generating synthetic positive samples from the anchor image. As shown in Fig. \ref{fig:overview}(a), we use a diffusion model to generate an additional positive sample $x_i^3$ and extract the embedding $z_i^3$ using the same encoder and projector employed in SimCLR. To utilize the information in the additional positive, we use a straightforward yet effective loss that minimizes the similarity of the positive pair $(x_i^1, x_i^3)$, $\mathcal{L}_{i}^{\prime}=||z_i^2-z_i^3||_2^2$. We use this loss because it is effective, decoupled with the original contrastive loss, and easy to control the magnitude with a simple weight factor $\lambda$. Therefore, the final loss function of CLSP is defined as:
\begin{equation}
\label{equation:l}
    \mathcal{L} = \sum_{i \in [1,N], k \in \{1,2\}}\mathcal{L}_{i,k} + \lambda \cdot \sum_{i \in [1, N]}||z_i^2-z_i^3||_2^2.
\end{equation}

\noindent\textbf{Hard Positive Generation with Diffusion Model.} Numerous studies have highlighted the exceptional image-generation capabilities of diffusion models. SOTA diffusion models are typically trained with semantic class information or text information to control the generated contents \cite{dhariwal2021diffusion, rombach2022high}. However, it is still unclear whether, given only an unlabeled anchor image, an unconditional diffusion model can generate diverse images that are semantically similar to the anchor image. Nor do we know the potential benefits of such synthetic images from diffusion models in the context of contrastive learning. In this section, we introduce a way to generate hard positives with a pre-trained diffusion model and demonstrate their efficacy in improving contrastive learning.

Our idea is based on the empirical observation that a well-trained unconditional diffusion model has already learned linear separable features in its intermediate layers. In Fig. \ref{fig:overview}(b), we visualize the t-SNE plot of the intermediate features extracted from an unconditional diffusion model trained on the CIFAR10 dataset. We can see that these features are well-clustered, indicating the model's capacity to map similar instances to proximate representations. Similar observations have also been reported in \cite{xiang2023denoising, yang2023diffusion}. If we mask out the intermediate features (set them to 0) during random sampling, the generated images will miss the semantic instance and only contain random backgrounds (refer to Fig. \ref{fig:overview}(c)). This suggests that the semantic and background information are decoupled in unconditional diffusion models. As such, we can leverage feature interpolation on the features from random sampling $h$ and features extracted from the diffusion process of the anchor image $h_{anchor}$ to generate new images resembling the anchor image. The interpolation process can be defined as:
\begin{equation}
\label{equation:h}
    h = w\cdot h + (1-w)\cdot h_{anchor}
\end{equation}

Where $w$ is the interpolation weight controlling the similarity of generated samples to the anchor image. Specifically, $w=1$ means pure random sampling with no feature interpolation, and $w=0$ does not mean the generated images are identical to the anchor image because the background-related features remain random. Our approach can be viewed as a specialized form of strong data augmentation, aiming to enhance the feature diversity of positive samples and guide the model's attention toward semantically meaningful regions in the image.

Practically, generating new positives on the fly for each image during training is unfeasible because of the slow sampling speed of diffusion models. Therefore, we pre-generated a positive candidate set with $k (k\leq 8)$ samples for each anchor image in the training set and randomly sampled one as the additional positive for CLSP training. This makes the CLSP training speed 50$\%$ slower than SimCLR. While we only demonstrate the workflow on CLSP on SimCLR, our method can be extended to MoCo and other similar contrastive learning frameworks.

\section{Experiment}
\label{sec:experiment}
This section empirically evaluates the proposed CLSP method and compares it to other SSL frameworks. We summarize the experiments and analysis as follows: (1) The proposed method significantly improves previous baselines on several benchmark datasets using linear evaluation protocol. (2) Our findings indicate that CLSP consistently outperforms other baseline methods in both semi-supervised and transfer learning scenarios, providing evidence for the generality of the learned representations. (3) To offer insights into the enhanced representation quality achieved by our proposed method, we conduct a detailed analysis of the similarity distribution of contrastive pairs.
All experiments are conducted with 4 Nvidia A10 GPUs on a single machine.

\subsection{Implementation Details}

\noindent \textbf{Dataset and Training Settings}
We evaluate the proposed approach across 4 datasets, including CIFAR10 \cite{krizhevsky2009learning}, CIFAR100 \cite{krizhevsky2009learning}, STL10 \cite{coates2011analysis}, and ImageNet100. ImageNet100, a subset of ImageNet1k \cite{deng2009imagenet} is frequently employed in various contrastive learning studies \cite{yeh2022decoupled, wu2023synthetic}. Unless otherwise specified, we use ResNet-18 as the backbone. A 2-layer MLP with dimensions [4096, 256] is used as the projector to project the representations. We implement two versions of our method CLSP-SimCLR and CLSP-MoCoV2. For CIFAR10, CIFAR100, and STL10, we trained the model for 1000 epochs with a batch size of 1024 for CLSP-SimCLR and 512 for CLSP-MoCoV2. We use SGD as the optimizer and a warmup cosine learning rate scheduler \cite{grill2020bootstrap}. The pre-generated candidate set size is 8, and the queue size for CLSP-MoCoV2 is 32768. Temperature $\tau$ is set to 0.2. We resize the image size to $64\times64$ for STL10 because it is easier to train an unconditional diffusion model on small image size. Only the \textit{unlabeld} set in STL10 is used for model pre-training. We pre-trained the unconditional diffusion model following the setting in \cite{ho2020denoising}. For the ImageNet100 dataset, we trained ResNet-50 for 200 epochs with a batch size of 256 to be consistent with some of the baselines. The pre-generated candidate set size is 4, and the CLSP-MoCoV2 queue size is 65536. The synthetic positives are generated with the pre-trained unconditional diffusion model provided in \cite{dhariwal2021diffusion}. The $\lambda$ for additional loss is set to 1.0 and 0.5 for CLSP-SimCLR and CLSP-MoCoV2, respectively.

\noindent \textbf{Baselines} We select representative baseline methods from \cite{JMLR:v23:21-1155}. Besides, we add some new baselines that incorporate new positives including (1) CL-GAN \cite{wu2023synthetic}, in which a GAN is jointly trained with the contrastive model to generate hard positives. (2) SimCLR-FT and MoCoV2-FT \cite{zhu2021improving}, feature transformation is used on embeddings to increase the hardness of positive pairs during training. (3) CLSP-noaug, we replace one of the augmented views in standard SimCLR with our synthetic positive to test the importance of the positive generated by naive data augmentation. (4) SimCLR-aug and MoCoV2-aug, we replace our additional synthetic positive in CLSP with a positive generated with naive data augmentation to test whether the benefit comes from the synthetic hard positive or the additional computation cost.

\subsection{Linear and kNN Evaluations}
We first compare our approach with current SOTA SSL techniques using linear and kNN evaluation protocols. In linear evaluation, a linear classifier is trained on the frozen encoder trained with SSL, and the test top1 accuracy is reported. Following the setting in \cite{chen2020simple}, we use SGD as the optimizer with no weight decay to train the linear classifier for 100 epochs. Only random cropping and horizontal flipping are used as data augmentation. For kNN evaluation, we use the cosine distance function, and the temperature is set to 0.07. We search for different $k$ values from $\{10,20,50,100,200\}$ and report the best result.

\begin{table}[t]
\centering
\setlength\tabcolsep{11.0pt}
\caption{Linear and kNN evaluation results on CIFAR10 and CIFAR100. Some baseline results are extracted from the solo-learn library \cite{JMLR:v23:21-1155}. All are based on ResNet-18.}
\begin{tabular}{lccc|cc}
\toprule
        & & \multicolumn{2}{c}{CIFAR10} & \multicolumn{2}{c}{CIFAR100} \\
        & Epoch & Linear & kNN & Linear & kNN  \\ \hline
Barlow Twins & 1000     & 92.10  & 90.60 & 70.90 & 66.40         \\
BYOL    & 1000 & 92.58  & 90.78 & 70.46 & 67.22        \\
DINO    & 1000 & 89.52  & 88.40 & 66.76 & 62.07         \\
MoCoV2  & 1000 & 92.94  & 91.45 & 69.89 & 66.91        \\
MoCoV3  & 1000 & 93.10  & 91.00 & 68.83 & 65.07           \\
NNCLR   & 1000 & 91.88  & 90.61 & 69.62 & 65.98        \\
SimCLR  & 1000 & 91.47  & 88.57 & 65.78 & 62.69        \\
SupCon  & 1000 & 93.82  & 92.60 & 70.38 & 69.28        \\
SwAV    & 1000 & 89.17  & 87.42 & 64.88 & 60.65        \\
VIbCReg & 1000 & 91.18  & 89.25 & 67.37 & 63.16        \\
VICReg  & 1000 & 92.07  & 89.97 & 68.54 & 65.09        \\
All4One & 1000 & 93.24 & 91.38  & \textbf{72.17} & 67.83      \\
DCL \cite{yeh2022decoupled} & 200 & 85.90 & 84.40 & 58.90 & 54.30 \\
CL-GAN \cite{wu2023synthetic} & 300 & 92.94 & - & 67.41 & - \\
SimCLR-FT \cite{zhu2021improving} & 1000 & 91.77 & 91.45 & 66.43 & 65.07 \\
MoCoV2-FT \cite{zhu2021improving} & 1000 & 92.58 & 92.60 & 69.40 & 68.79 \\
CLSP-noaug & 1000 & 92.86 & 92.46 & 54.46 & 53.53 \\ 
SimCLR-aug & 1000 & 91.96 & 91.44 & 68.30 & 66.75 \\
MoCoV2-aug & 1000 & 93.04 & 92.82 & 69.13 & 68.21 \\ \hline
CLSP-SimCLR & 300 & 93.10 & 92.44 & 67.94 & 66.04 \\
CLSP-SimCLR & 1000 & 94.37 & 93.59 & 72.01 & \textbf{70.19} \\
CLSP-MoCoV2 & 1000 & \textbf{94.41} & \textbf{93.90} & 71.76 & 69.75 \\
\bottomrule
\end{tabular}
\vspace{-15pt}
\label{tab:linear_cifar}
\end{table}

We report the linear and kNN classification results on CIFAR datasets for all methods in Table \ref{tab:linear_cifar}. As we can see, both our CLSP-SimCLR and CLSP-MoCoV2 outperform the previous methods in both Linear and kNN evaluations on CIFAR10, even better than SupCon which includes label information in the pre-training stage. Compared to SimCLR and MoCoV2, our improved versions gain 2.90\% and 1.47\% linear accuracy improvement on CIFAR10, along with 6.23\% and 1.87\% gains on CIFAR100. Notice that on CIFAR100, our CLSP-SimCLR slightly underperforms the All4One baseline by 0.16\%, the reason could be their reported number is evaluated online during the pre-training when extensive data augmentation is used whereas we only use random crop and random flipping. When evaluated offline in the same setting, our method performs the best among all the baselines, as shown in CIFAR100 kNN results in Table \ref{tab:linear_cifar}. Comparing the results of SimCLR, CLSP-noaug, and CLSP-SimCLR, we can see that the integration of harder positives can indeed improve performance even without naive augmented positive pairs on CIFAR10, but the conclusion does not hold on CIFAR100. We believe this is because the instance discrimination task for CIFAR100 is more challenging than CIFAR10, naive data augmentated pairs are needed to stabilize the training. The optimal performance in representation learning is achieved by combining both augmented positive pairs and synthetic positives. Furthermore, SimCLR-aug improves the linear classification accuracy by 0.49\% and 2.52\% over SimCLR on CIFAR10 and CIFAR100. This means that introducing new augmented views does improve the model, but the improvement is marginal compared to CLSP-SimCLR, which proves that the synthetic hard positives benefit contrastive learning more than the augmentations of real data.

In Table \ref{tab:linear_stl10}, we compared the linear evaluation results of our method with the selected baselines on the STL10 and ImageNet100 datasets. We use ResNet-18 on STL10 and ResNet-50 on ImageNet100 to match the setting of previous works \cite{yeh2022decoupled, wu2023synthetic}. Considering the computational resources for ImageNet100, we pre-train the model for 200 epochs and perform the linear evaluation for 100 epochs. From the results, we can see that the proposed CLSP method outperforms existing SSL baselines. Specifically, substantial improvements of 3.11\% and 3.87\% over the original SimCLR framework are observed on both datasets. This shows that the proposed can be scaled to large-scale datasets like ImageNet100.

\begin{table}[t]
\centering
\setlength\tabcolsep{6.0pt}
\caption{Linear evaluation results on STL10 and ImageNet100. All methods use ResNet-18 on STL10, and ResNet-50 on ImageNet100.}
\begin{tabular}{lcc|cc}
\toprule
        & \multicolumn{2}{c}{STL10} & \multicolumn{2}{c}{ImageNet100} \\
        & Epoch & Linear (Top1) & Epoch & Linear (Top1) \\ \hline
BYOL    & 1000 & 92.11 & 200 &  75.80         \\
MoCoV2  & 1000 & 92.20  & 200 &  78.00        \\
NNCLR   & 1000 & 90.78 & 200 & 77.43      \\
SimCLR  & 1000 & 90.63 & 200 &  75.75        \\
All4One & 1000 & 92.21 & 200 & 79.09    \\
CL-GAN \cite{wu2023synthetic} & - & - & 200 & 78.85 \\
SimCLR-FT \cite{zhu2021improving} & 1000 & 90.95 & 200 & 76.27 \\
SimCLR-aug & 1000 & 91.34 & 200 & 78.32 \\ \hline
CLSP-SimCLR & 1000 & \textbf{93.74} & 200 & \textbf{79.62} \\
CLSP-MoCoV2 & 1000 & 93.69 & 200 & 79.11 \\
\bottomrule
\end{tabular}
\label{tab:linear_stl10}
\end{table}

\begin{table}[t]
\centering
\setlength\tabcolsep{6.0pt}
\caption{Semi-supervised learning results. Top-1 accuracy is reported on training the classifier on the fixed ResNet-18 encoder with 1\% and 10\% labeled data.}
\begin{tabular}{lcc|cc|cc}
\toprule
        & \multicolumn{2}{c}{CIFAR10} & \multicolumn{2}{c}{CIFAR100} & \multicolumn{2}{c}{STL10} \\
        & 1\% & 10\% & 1\% & 10\% & 1\% & 10\% \\ \hline
SimCLR  & 84.62 & 87.20 & 39.00 & 55.94 & 69.63 & 88.05 \\
MoCoV2  & 88.75 & 91.19 & 44.52 & 61.37 & 72.60 & 90.52 \\
BYOL    & 86.01 & 90.24 & 43.54 & 61.30 & 72.41 & 88.85 \\
NNCLR   & 85.55 & 88.80 & 39.69 & 60.09 & 69.96 & 88.96  \\
All4One & 87.70 & 90.24 & 44.18 & 62.31 & 73.60 & 89.06  \\
SimCLR-FT & 87.71 & 90.69 & 44.06 & 62.05 & 69.85 & 88.13 \\
SimCLR-aug & 87.89 & 90.80 & 42.73 & 60.07 & 69.74 & 88.85 \\  \hline
CLSP-SimCLR & \cellcolor{piggypink}90.57 & \cellcolor{piggypink}93.47 & \cellcolor{piggypink}48.08 & \cellcolor{piggypink}65.01 & \cellcolor{piggypink}74.20 & \cellcolor{piggypink}92.05 \\
CLSP-MoCoV2 & \cellcolor{piggypink}91.13 & \cellcolor{piggypink}93.53 & \cellcolor{piggypink}46.84 & \cellcolor{piggypink}63.61 & \cellcolor{piggypink}76.64 & \cellcolor{piggypink}92.16 \\
\bottomrule
\end{tabular}
\label{tab:semi}
\end{table}

\subsection{Semi-supervised and Transfer Learning}

We evaluate the effectiveness of our method in a semi-supervised setting on CIFAR10, CIFAR100, and STL10 with 1\%, 10\% subsets following the standard evaluation protocol \cite{chen2020simple, grill2020bootstrap}. The results are presented in Table \ref{tab:semi}. We can see that our method outperforms all the SOTA baseline methods on semi-supervised learning in all settings, this proves the robust generalization capacity of CLSP features, particularly when dealing with limited label availability.

Moreover, we show that the representations learned using CLSP are effective for transfer learning across multiple downstream classification tasks. We follow the linear evaluation setup described in \cite{Estepa_2023_ICCV}. The datasets used in this benchmark are as follows: CIFAR10, CIFAR100, Oxford-IIIT-Pets \cite{parkhi2012cats}, Oxford-Flowers \cite{nilsback2008automated}, DTD \cite{cimpoi2014describing}, Caltech-101 \cite{fei2004learning}, Food101 \cite{bossard2014food}, and Cars \cite{krause20133d}. For all datasets, we freeze the encoder pre-trained on the source dataset and train a single linear classifier on top of it for 100 epochs on the target training set while performing a grid search on the learning rate to find the optimal one. Then, we evaluate the performance on the target test set. The results are shown in Table \ref{tab:transfer} and Table \ref{tab:transfer_imagenet100}. As can be seen, CLSP outperforms the baselines in most downstream datasets, further validating the superior generalization capabilities of the learned representations.

\begin{table}[t]
\centering
\caption{Transfer learning to downstream tasks (Top1 linear classification accuracy).}
\setlength\tabcolsep{2.0pt}
\resizebox{1.0\columnwidth}{!}{
\begin{tabular}{lccccccccc}
\toprule
Source        & CIFAR10 & CIFAR100 & \multicolumn{6}{c}{STL10} \\ \cmidrule(lr){2-2}  \cmidrule(lr){3-3} \cmidrule(lr){4-9} 
Target & CIFAR100 & CIFAR10 & CIFAR10 & CIFAR100 & Pets & Flowers & DTD & Caltech101 \\ \hline
SimCLR  & 47.97 & 73.81 & 81.58 & 52.79 & 53.97 & 53.23 & 45.74 & 73.42 \\
MoCoV2  & 54.90 & 78.57 & 83.39 & 55.41 & 56.37 & 54.32 & 46.23 & 76.13 \\
BYOL    & 53.62 & 78.69 & 82.88 & 55.34 & 55.93 & 57.81 & 45.59 & 78.57 \\
NNCLR   & 53.27 & 78.79 & 81.60 & 53.94 & 56.31 & \textbf{61.93} & 43.51 & 78.24 \\
All4One & 56.17 & 79.38 & 83.33 & 55.49 & 56.65 & 57.52 & 46.13 & 76.62  \\
SimCLR-FT & 55.36 & 80.89 & 81.62 & 53.89 & 54.70 & 51.75 & 42.71 & 73.58 \\
SimCLR-aug & 54.42 & 79.78 & 81.83 & 54.57 & 56.66 & 56.14 & 45.37 & 75.16 \\ \hline
CLSP-SimCLR & \textbf{61.64} & \textbf{83.14} & 84.98 & 56.83 & \textbf{56.75} & 57.52 & \textbf{46.76} & \textbf{80.01} \\
CLSP-MoCoV2 & 61.19 & 83.13 & \textbf{86.58} & \textbf{57.97} & 55.82 & 50.20 & 44.68 & 79.42 \\
\bottomrule
\end{tabular}
}
\label{tab:transfer}
\end{table}

\begin{table}[t]
\centering
\caption{ImageNet100-pretrained transfers to downstream datasets.}
\setlength\tabcolsep{4.0pt}
\resizebox{1.0\columnwidth}{!}{
\begin{tabular}{lcccccccc}
\toprule
 & CIFAR10 & CIFAR100 & Pets & Flowers & DTD & Caltech101 & Food101 & Cars \\ \hline
All4One & 86.58 & 65.42 & 71.25 & 72.21 & 59.04 & 84.93 & \textbf{63.30} & \textbf{33.14} \\
CLSP-SimCLR & \textbf{87.36} & \textbf{66.50} & \textbf{73.97} & \textbf{72.65} & \textbf{59.95} & \textbf{85.24} & 59.87 & 32.81  \\
\bottomrule
\end{tabular}
}
\label{tab:transfer_imagenet100}
\end{table}

\subsection{Similarity Distribution of Contrastive Pairs}

\begin{figure}[t!]
	\centering
	\includegraphics[width=0.9\linewidth]{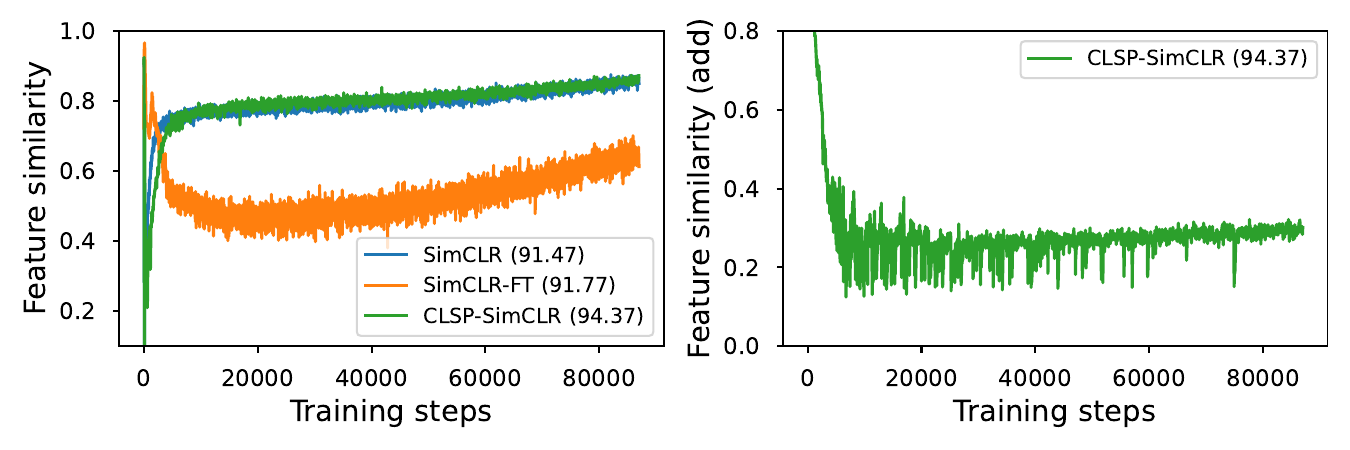}
	\vspace{-10pt}
	\caption{Feature similarity of original positive pairs (left) and additional positive pairs (right) on CIFAR10. The original positive pair is the two augmented views, and the additional positive pair is one of the augmented views with the synthetic positive.}
	\label{fig:feature_similarity}
\end{figure}

The similarity of positive pairs can reflect the performance of a contrastive learning model \cite{zhu2021improving}. A small positive similarity means this positive pair carries a large view variance of the sample, which can help the model learn more view-invariant features and improve the performance. \cite{zhu2021improving} controls the hardness of positive pairs by changing the momentum and doing feature extrapolation, they modify the original features so that the positives become harder to train. However, our method does not explicitly change the original positive features. Instead, we add a new positive pair which is much harder to train than the original pair. In Fig. \ref{fig:feature_similarity}, we visualize the distribution of the original positive and the additional positive from different methods. It can be seen that SimCLR-FT alters the original similarity and makes it smaller while our synthetic positive pairs help the model learn the original positive better (larger feature similarity than SimCLR), which explains the observed improvements in linear and transfer learning evaluation.

\begin{table}[t]
\centering
\setlength\tabcolsep{4.0pt}
\caption{Linear evaluation of CLSP-SimCLR under different training epochs and backbone size on CIFAR10 and CIFAR100.}
\begin{tabular}{llccc}
\toprule
Method  & Backbone & Epoch & CIFAR10 & CIFAR100 \\ \hline
SimCLR  & ResNet-18 & 2000 & 92.1 & 68.1     \\
SimCLR  & ResNet-50 \cite{chen2020simple}  & 1000 & 94.0 & 70.7 \\
SimCLR  & ResNet-101-SK \cite{chen2020big} & 1000 & 96.4 & 74.3 \\
CLSP-SimCLR & ResNet-18 & 2000 & 94.7 & 72.1      \\
CLSP-SimCLR & ResNet-50 & 1000 & 95.8 & 73.7 \\
CLSP-SimCLR & ResNet-101-SK & 1000 & 96.9 & 75.4 \\
\bottomrule
\end{tabular}
\vspace{-12pt}
\label{tab:model_size}
\end{table}

\section{Ablations}

In this section, we perform extensive ablations on the hyperparameter of the proposed CLSP framework on CIFAR datasets. We show that CLSP is robust to the model size and training epochs. By seeking the optimal settings hyperparameters like candidate set size and interpolation weight, we demonstrate the best practice for CLSP.

\noindent \textbf{Model Size and Training Time} In Section \ref{sec:experiment}, we evaluate the performance of CLSP with ResNet-18 backbone on various datasets. To test if the improvements still hold with larger model sizes and longer training, we evaluate the CLSP with doubled training iterations and larger backbones. The results are presented in Table \ref{tab:model_size}. It can be seen that with larger training epochs, the accuracy gain over SimCLR is still significant. In addition, CLSP-SimCLR outperforms the SimCLR when the encoder backbone becomes larger. Notice that the accuracy of CLSP-SimCLR ResNet-18 on CIFAR10 and CIFAR100 datasets (94.4, 72.0) surpasses the baseline ResNet-50 (94.0, 70.7) under the same training epochs.

\noindent \textbf{Positive Candidate Set Size} Ideally, we would like to generate synthetic positives on the fly to maximize the feature diversity of the generated samples. However, in practice, we need to pre-generate $k$ positives for each image in the dataset, where $k$ is the positive candidate set size. If $k$ is extremely small (e.g., $k=1$), the model optimizes the same additional positive pair throughout the entire training process, which is suboptimal. Conversely, if $k$ is too large, we need more space to store the additional positives, introducing additional overhead. In table \ref{tab:pool_size}, we evaluate the linear evaluation performance of CLSP-SimCLR under different $k$. The results show that the performance first improves when $k$ becomes larger and then stops increasing when $k$ reaches a certain threshold.

\begin{table}[t]
\centering
\setlength\tabcolsep{8.0pt}
\caption{Linear evaluation of CLSP-SimCLR under different positive candidate set sizes on CIFAR10 and CIFAR100.}
\begin{tabular}{ccccccc}
\toprule
$k$  & 1 & 2 & 4 & 8 & 16 & 32 \\ \hline
CIFAR10  & 92.24 & 93.29 & 93.95 & 94.37 & 94.38 & 94.33     \\
CIFAR100 & 65.84 & 68.33 & 70.65 & 72.01 & 71.92 & 72.04  \\
\bottomrule
\end{tabular}
\label{tab:pool_size}
\end{table}

\begin{figure}[t!]
	\centering
	\vspace{-3pt}
	\includegraphics[width=1.0\linewidth]{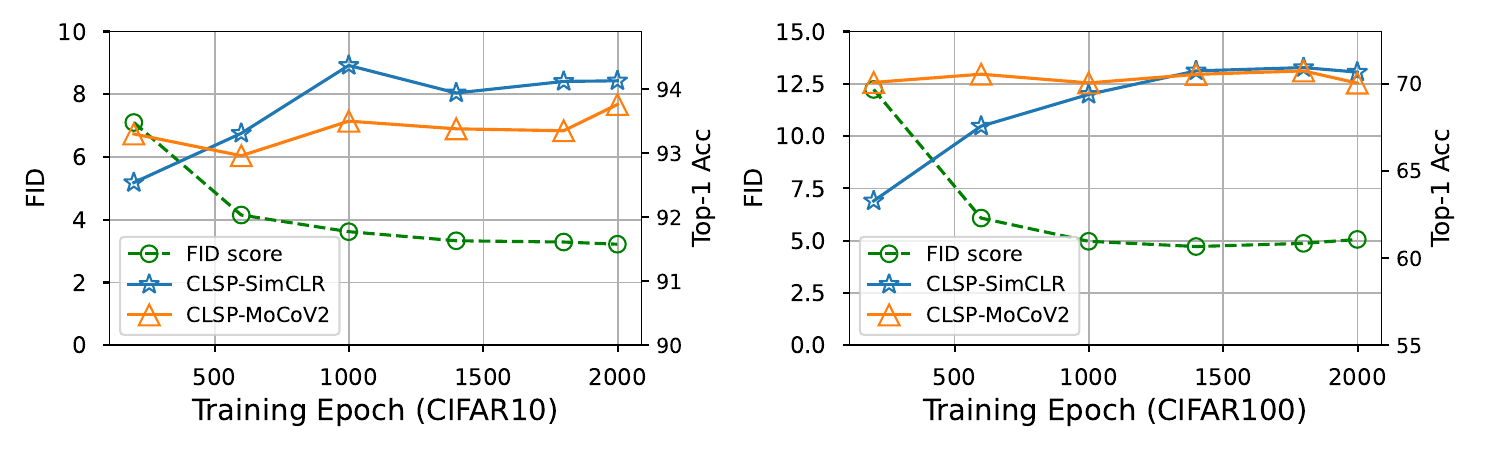}
	\vspace{-15pt}
	\caption{The correlation of image generation quality with CLSP-SimCLR and CLSP-MoCoV2 linear evaluation performance on CIFAR10 and CIFAR100 datasets.}
	\label{fig:fid_acc}
\end{figure}

\noindent \textbf{Generation Quality of Diffusion Model} One of the main reasons we use diffusion models is their superior generation quality. A natural question one might ask is what if the diffusion model is not so good (e.g., not trained to the optimal state), how does the generation quality affect the final SSL performance? To answer this question, we use the diffusion model checkpoints from different training stages to generate synthetic positives under the same setting. The interpolation weight $w=0.1$, the candidate set $k=8$. The correlation of diffusion model quality (FID score) and SSL linear evaluation accuracy is shown in Fig. \ref{fig:fid_acc}. We can observe that the SSL linear evaluation accuracy becomes larger when the diffusion model becomes better. When the diffusion model training epoch is larger than 1000 epochs, the SSL accuracy tends to remain constant.

\begin{figure}[t!]
	\centering
        \begin{subfigure}[b]{0.5\linewidth}
		\includegraphics[width=\linewidth]{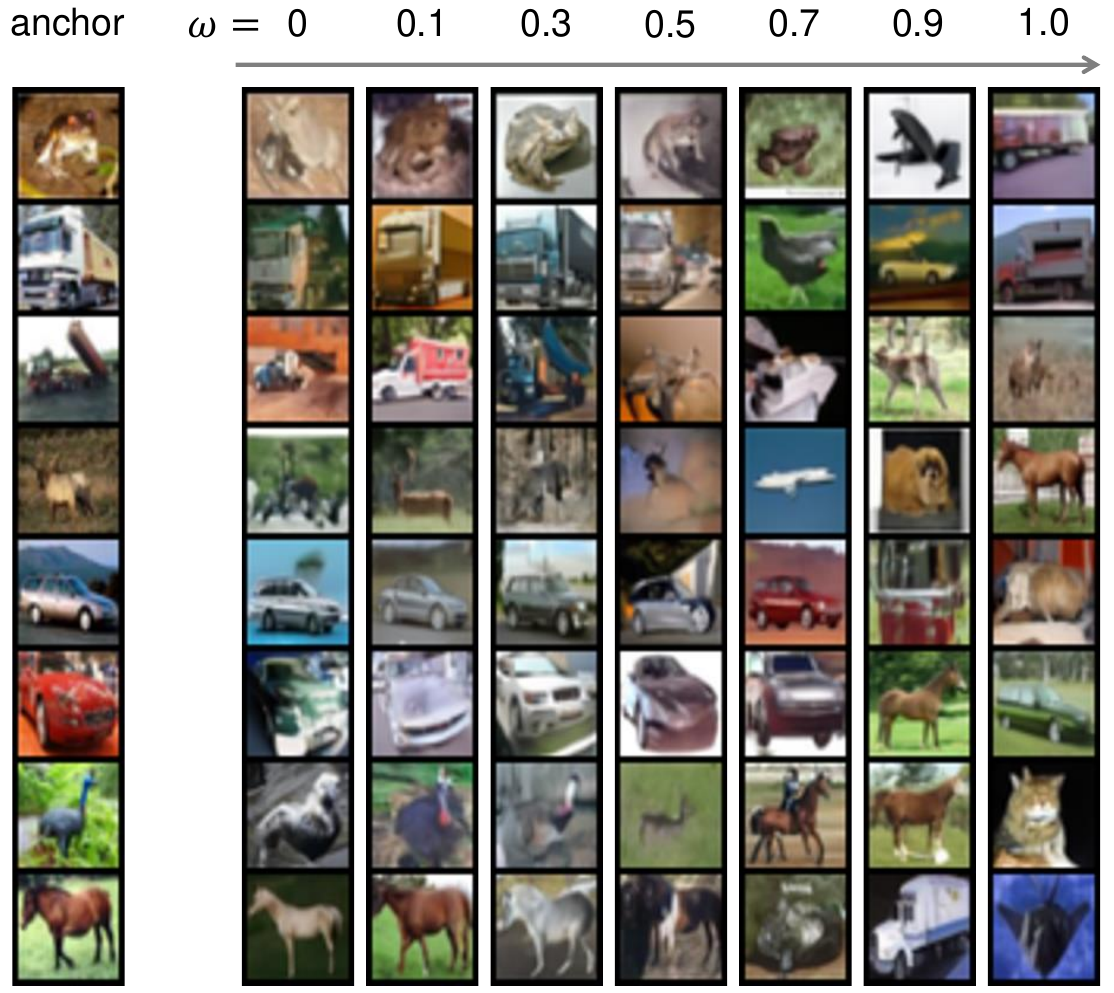}
		\caption{}
		\label{fig:}
	\end{subfigure}
	\hspace{2pt}
	\begin{subfigure}[b]{0.46\linewidth}
		\includegraphics[width=\linewidth]{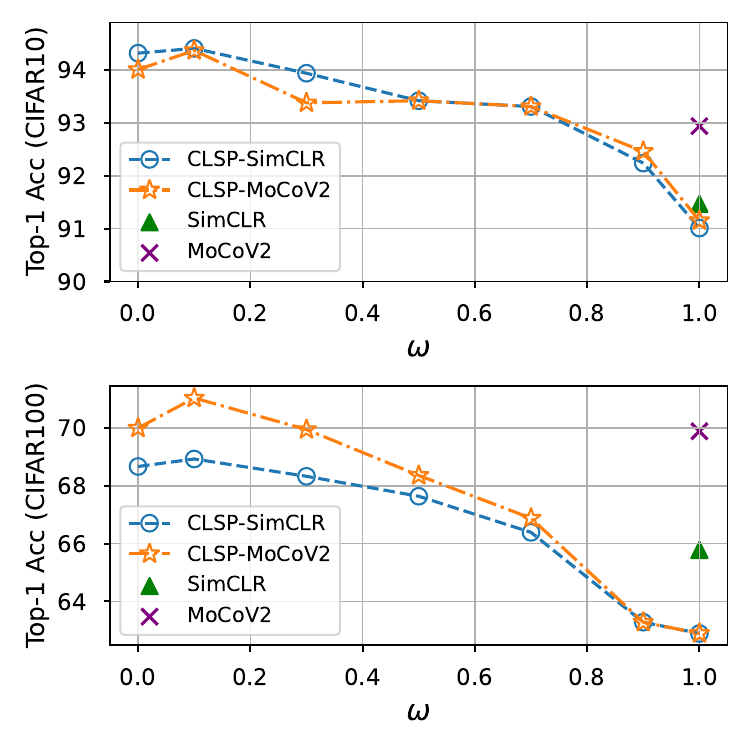}
		\vspace{-16pt}
		\caption{}
		\label{fig:}
	\end{subfigure}
	\vspace{-5pt}
	\caption{(a) Generated images under different feature interpolation weights. (b) The correlation of linear classification accuracy with $w$ on CIFAR10 and CIFAR100 datasets}
	\label{fig:acc_w}
\end{figure}

\noindent \textbf{Feature Interpolation Weight for Positive Generation} In Equation \ref{equation:h}, the feature interpolation weight $w$ controls the degree of randomness in the semantic features during the generation process. A higher value of $w$ results in increased randomness and diversity in generated positives. However, it also increases the probability of false positives. We visualize the generated images under different interpolation weights in Fig. \ref{fig:acc_w}(a), along with the correlation of linear evaluation accuracy with $w$ in Fig. \ref{fig:acc_w}(b). It can be seen that large interpolation weights usually decrease the linear classification accuracy. Empirically, $w=0.1$ could be a good starting point which usually leads to promising results.

\noindent \textbf{Additional Loss Weight.} In Equation \ref{equation:l}, the additional loss weight $\lambda$ controls the magnitude of optimizing the additional positive pair. If $\lambda$ is too small, the benefit of synthetic positive might be gone. If $\lambda$ is too large, the model will focus too much on hard positives, making it easier to collapse. Table \ref{tab:ap_loss_weight} shows the linear evaluation results of CLSP-SimCLR and CLSP-MoCoV2 with different $\lambda$.

\begin{table}[t]
\centering
\setlength\tabcolsep{4.0pt}
\caption{Linear evaluation results of CLSP-SimCLR under different additional loss weight $\lambda$ on CIFAR10 and CIFAR100.}
\begin{tabular}{cccccccccc}
\toprule
$\lambda$  & 0.01 & 0.2 & 0.4 & 0.6 & 0.8 & 1.0 & 1.2 & 1.4 & 1.6 \\ \hline
CIFAR10  & 91.80 & 91.97 & 93.16 & 93.84 & 94.23 & 94.37 & 94.27 & 94.21 & 94.59    \\
CIFAR100 & 65.19 & 68.08 & 70.68 & 71.36 & 71.81 & 72.01 & 70.44 & 70.00 & 69.07  \\
\bottomrule
\end{tabular}
\label{tab:ap_loss_weight}
\vspace{-10pt}
\end{table}

\noindent \textbf{Additional Positive Number.} In our earlier discussions, we focused on the incorporation of a single additional positive branch within the standard SimCLR framework. We demonstrated that adding just one additional positive, whether generated through data augmentation or a diffusion model, enhances the performance of standard SimCLR. Here, we explore whether introducing more than one additional positive could further elevate the representation capacity. Table \ref{tab:ap_number} presents the linear evaluation results of CLSP with 1, 2, and 4 additional positives for each anchor image on the CIFAR10 dataset. We can see that adding more positives does not further improve contrastive learning.

\begin{table}[t]
\centering
\setlength\tabcolsep{8.0pt}
\caption{Linear evaluation results of CLSP-SimCLR and CLSP-MoCoV2 on CIFAR10 dataset with different numbers of additional positives.}
\begin{tabular}{lccc}
\toprule
Additional Positive \#  & 1 & 2 & 4 \\ \hline
CLSP-SimCLR  & 94.37 & 94.31 & 94.35  \\
CLSP-MoCoV2 & 94.43 & 94.40 & 94.21  \\
\bottomrule
\end{tabular}
\label{tab:ap_number}
\end{table}

\section{Discussion}

\noindent \textbf{Alternative Positive Generation Methods.} While we demonstrated that using feature interpolation in diffusion sampling is an effective way to generate harder positives for contrastive learning, alternative methods exist for positive generation with diffusion models. For example, recent work RCG \cite{li2023self} proposes an approach where the diffusion model is conditioned on the embedding generated from the anchor image to generate high-quality images similar to the anchor image. However, this method requires a pre-trained encoder, usually trained with SSL, to generate embeddings as the condition. It remains unclear whether the generated positive could, in return,  benefit the original SSL performance. In addition, we could cluster the embeddings into a fixed number of clusters and use the cluster as the condition. The sampling will be similar to the class-conditioned diffusion model, and the generated images will be more diverse because they are not directly related to the original embedding. We leave the further discussion and comparison to the supplementary.

\noindent \textbf{Limitations.} Despite the promising results achieved by our CLSP method, we identify the following limitations. (1) \textbf{Computation and Memory Overhead}. The additional computation costs of CLSP include the diffusion model training and sampling, which could be huge if the training set is large. Because we pre-generate the additional positives into a candidate set, additional memory is needed to save the generated positives. Besides, one additional branch is added to SimCLR, so an extra 50\% computation overhead is needed. (2) \textbf{Increased Number of Hyperparameters} The positive generation part has two hyperparameters, including the feature interpolation weight and candidate set size. The CLSP loss introduces additional loss weight as an extra hyperparameter.

\section{Conclusion}
This paper introduces CLSP, a novel framework for contrastive representation learning with synthetic positives generated by a diffusion model. We show that using feature interpolation in the sampling process of an unconditional diffusion model can guide the model to generate diverse hard positives of the anchor image. By incorporating these synthetic hard positives in the contrastive learning process, the model can learn improved semantic features and achieve SOTA results on various benchmarks. In the future, we plan to extend CLSP to more complex datasets such as ImageNet1k and investigate its transfer learning performance on other computer vision tasks such as object detection.

\clearpage  

%
%
\bibliographystyle{splncs04}
\bibliography{main}

\cleardoublepage
\appendix
\counterwithin{figure}{section}
\counterwithin{table}{section}

\section{Appendix}

\subsection{Alternative Positive Generation Methods}

\begin{figure}[h]
	\centering
	\vspace{-10pt}
	\includegraphics[width=1.0\linewidth]{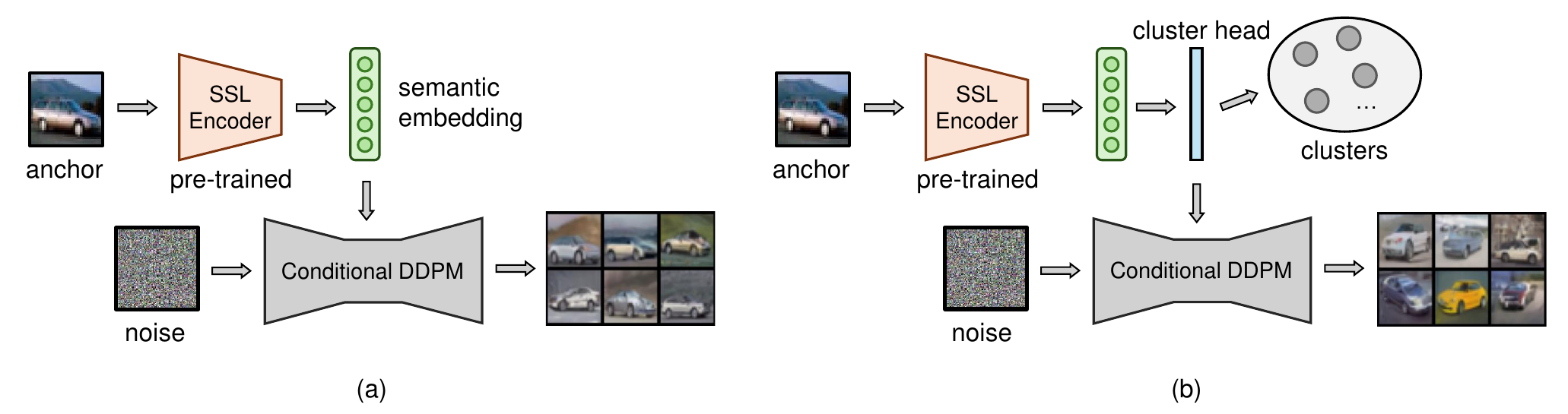}
	\vspace{-20pt}
	\caption{Alternative positive generation methods. (a) RCG, using a pre-trained SSL encoder to generate semantic embedding of the anchor image as the condition to guide diffusion sampling. (b) RCG-cluster, using an unsupervised cluster head to cluster the embeddings in RCG and then using the cluster output as the condition.}
 \label{fig:alternative_method}
\end{figure}

\begin{figure}[t]
	\centering
	\includegraphics[width=0.8\linewidth]{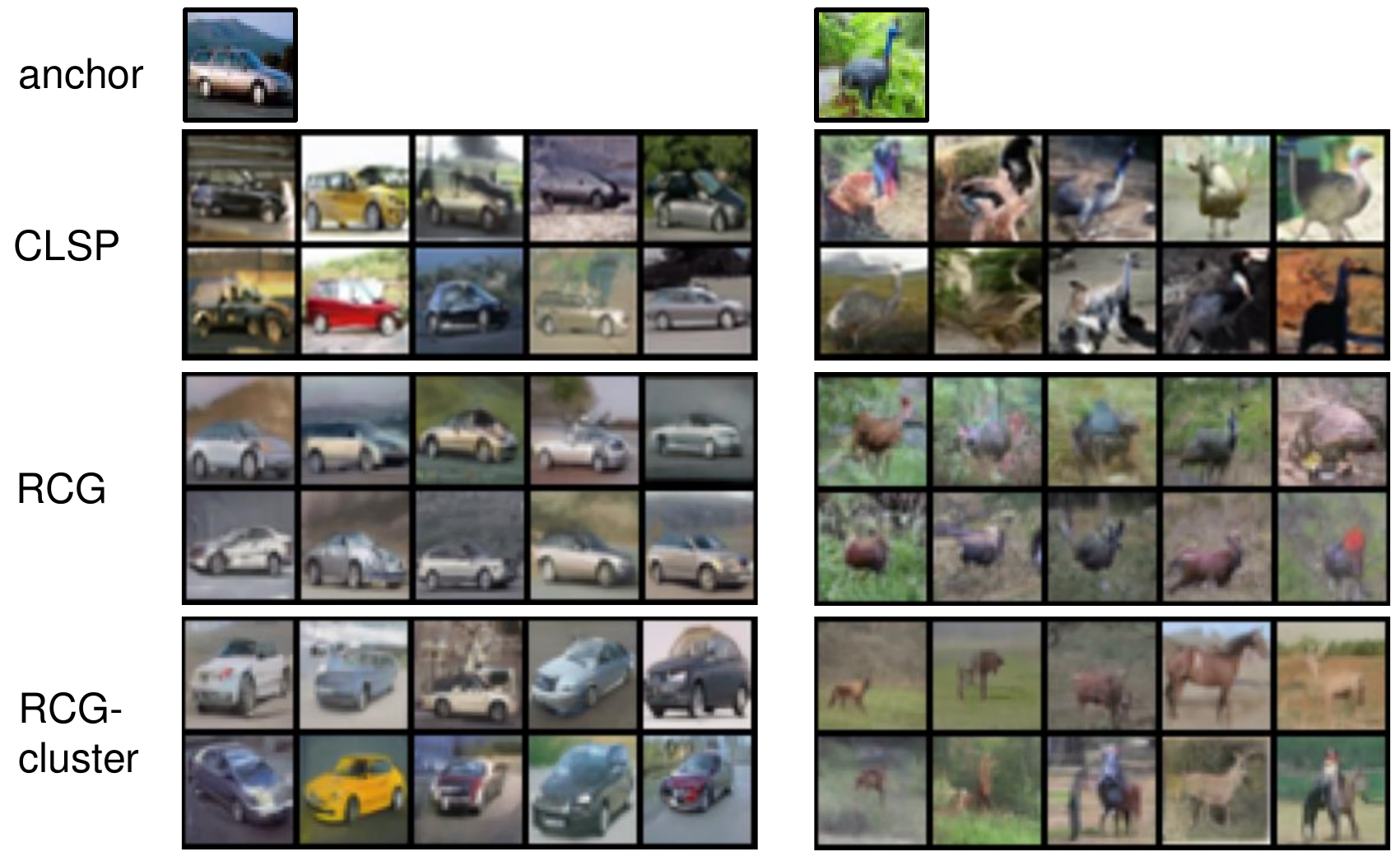}
	\caption{Visualization of the synthetic positives generated by different methods. Compared to CLSP, the positives generated by RCG are more similar to the anchor image and thus ``easier'' for contrastive learning. RCG-cluster brings more variance to the synthetic positives. However, it also introduces more false positives when the anchor image is not mapped to the right cluster.}
 \label{fig:alternative_method_example}
\end{figure}

In this section, we discuss two alternative positive generation methods using the diffusion model and compare them with our proposed CLSP. (1) Similar to the idea in \cite{li2023self}, we use an image encoder (pre-trained by any SSL method) to transfer the raw image distribution to a low-dimensional semantic embedding. Subsequently, we train a conditional diffusion model to map a noise distribution to the image distribution conditioned on the semantic embedding. This approach, termed RCG, is illustrated in Fig. \ref{fig:alternative_method}(a). Due to the influence of semantic embedding on the diffusion sampling process, the resulting synthetic images typically contain similar semantic content to that of the anchor image and can be used as additional positives. However, one of the key drawbacks of RCG is that the synthetic positives often closely resemble the anchor image, potentially diminishing the benefits of learning these "easy" positives compared to CLSP. As can be seen in Fig. \ref{fig:alternative_method_example}, RCG-generated images exhibit a higher visual similarity to the anchor images, with less variation in semantic content. (2) To mitigate the aforementioned limitation in RCG, we can use a cluster head (e.g., MLP) to cluster the semantic embeddings generated in RCG into distinct clusters and then use these clusters as the condition to guide diffusion sampling. We name this method RCG-cluster (Fig. \ref{fig:alternative_method}(b)). The idea is inspired by the class conditional diffusion model \cite{ho2021classifier}, where a pre-trained model can generate diverse images belonging to a specific class given its label. In RCG-cluster, the cluster head is trained using unsupervised learning techniques such as k-means, and the resulting clusters serve as conditions similar to those in the class conditional diffusion model. RCG-cluster has the potential to generate even more diverse positives than CLSP because the generated images are not explicitly correlated with the semantic embedding of the anchor image. However, it may introduce an increased risk of false positives when the anchor image is not mapped into the right cluster (Fig. \ref{fig:alternative_method_example}).

Table \ref{tab:alternative_methods_result} demonstrates the linear evaluation results of different positive generation methods on CIFAR datasets. For fast evaluation, we set the cluster size in the RCG-cluster to 10 and 100 for CIFAR10 and CIFAR100, respectively. We can see that our CLSP-SimCLR performs the best among these three methods on both CIFAR10 and CIFAR100. RCG-cluster performs better than RCG on CIFAR10 but worse on CIFAR100, the reason could be that CIFAR100 has more classes, which makes the cluster head less accurate and consequently yields more false positives to contrastive learning.

\begin{table}[t]
\centering
\setlength\tabcolsep{8.0pt}
\caption{Linear evaluation results of different positive generation methods on CIFAR datasets. All methods are based on the SimCLR framework.}
\begin{tabular}{lcc}
\toprule
Method  & CIFAR10 & CIFAR100 \\ \hline
RCG  & 92.41 & 67.90  \\
RCG-cluster & 92.65 & 66.20 \\
CLSP-SimCLR & 94.37 & 72.01 \\
\bottomrule
\end{tabular}
\label{tab:alternative_methods_result}
\end{table}

\subsection{Implementation Details}

\noindent \textbf{Diffusion Model.} We pre-train our unconditional diffusion models on CIFAR10, CIFAR100, and STL10 datasets. Following the setting in \cite{ho2020denoising}, we use U-Net with an initial filter size of 128 as the backbone. We use a linear $\beta$ scheduler from $\beta_1=1e-4$ to $\beta_T=0.02$, $T=1000$. Self-attention is used at the 16$\times$ 16 feature map resolution. Dropout ratio is set to 0.1. We train the diffusion model for 2000 epochs with Adam optimizer and a fixed learning rate 0.0002. Only random horizontal flipping is used as the data augmentation. For the STL10 dataset, we resize the image to 64 $\times$ 64 and only use the \textit{unlabeled} split for training. We use DDIM to speed up the diffusion sampling process and the sampling timestamp is set to 200. For CIFAR10, CIFAR100, and STL10 datasets, we use the features from the last layer of the U-Net encoder for feature interpolation. For ImageNet100, we use the features from the last 4 layers of the encoder for feature interpolation.

\noindent \textbf{Self-supervised Pre-training.} We use ResNet-18 as the default backbone for all experiments. On CIFAR10 and CIFAR100 datasets, we replace the first $7\times 7$ Conv of stride 2 with $3\times 3$ Conv of stride 1 and remove the maxpooling layer. For STL10, we just replace the first convolutional layer as in CIFAR datasets and keep the maxpooling layer. The batch size is 1024 for SimCLR-based approaches and 512 for MoCo-based approaches on CIFAR10, CIFAR100, and STL10. The batch size is 256 for all methods on ImageNet100. We use SGD as the optimizer with a weight decay of 0.0001 and momentum of 0.9. The initial learning rate is 0.3 with a cosine decay schedule, and linear warmup is used for the first 10 epochs. The temperature $\tau$ is 0.2 for all methods. We use standard data augmentations for both anchor images and synthetic images, including random cropping and resizing, random flipping, color distortion, grayscale, and polarization.

\noindent \textbf{Linear Evaluation.} Following the similar offline linear evaluation procedure in \cite{chen2020simple}, we freeze the encoder and train a linear classifier using an SGD optimizer without weight decay for 100 epochs. Only random cropping and random flipping are used as data augmentation. The initial learning rate is 10 and reduced to 1.0 and 0.1 at 60 and 80 epochs.

\begin{figure}[t!]
	\centering
        \begin{subfigure}[b]{0.48\linewidth}
		\includegraphics[width=\linewidth]{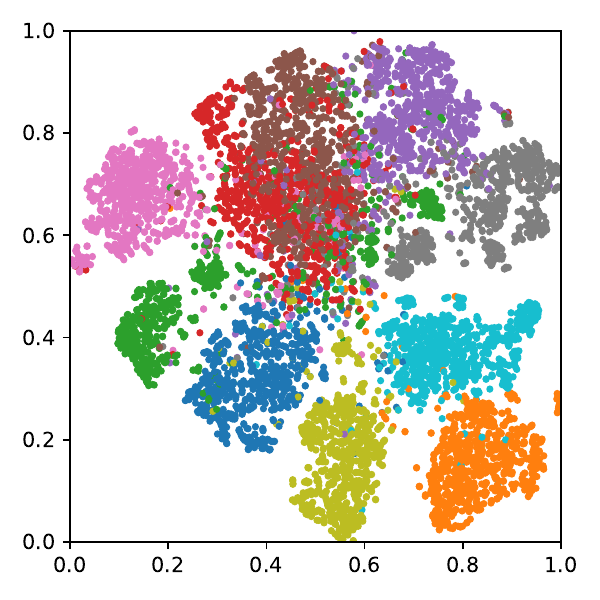}
		\caption{SimCLR}
		\label{fig:}
	\end{subfigure}
	\hspace{2pt}
	\begin{subfigure}[b]{0.48\linewidth}
		\includegraphics[width=\linewidth]{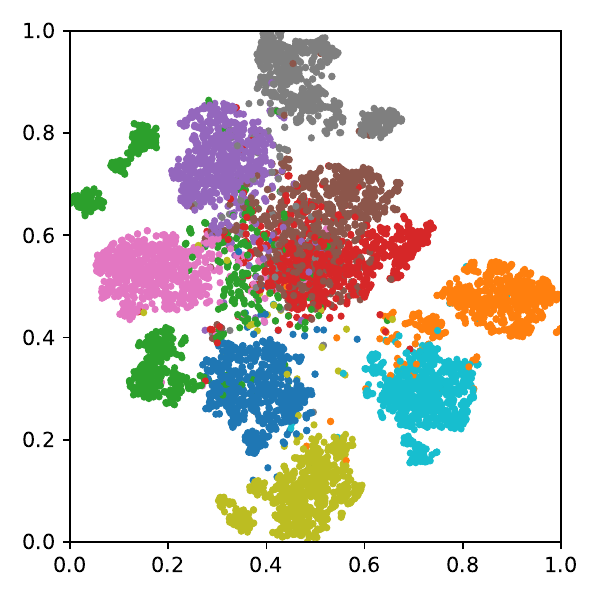}
		\caption{CLSP-SimCLR}
		\label{fig:}
	\end{subfigure}
	\caption{t-SNE visualization of representations learned by SimCLR and CLSP-SimCLR. Each color represents the representation of a specific class.}
 \vspace{-12pt}
	\label{fig:tsne_plot}
\end{figure}

\subsection{Visualization of Representations}
We compare the representations learned by our proposed CLSP-SimCLR and standard SimCLR using t-SNE visualization in Fig. \ref{fig:tsne_plot}. It can be seen that SimCLR has more overlapped representations among different classes. Such overlapped representations are non-discriminative and offer less information for the downstream tasks. However, the clusters learned by the proposed CLSP-SimCLR are denser and separable, meaning that the SSL model learns better discriminative features of different classes and provides more information for downstream tasks.

\subsection{Impact of Batch Size}

\begin{figure}[t]
	\centering
	\includegraphics[width=1.0\linewidth]{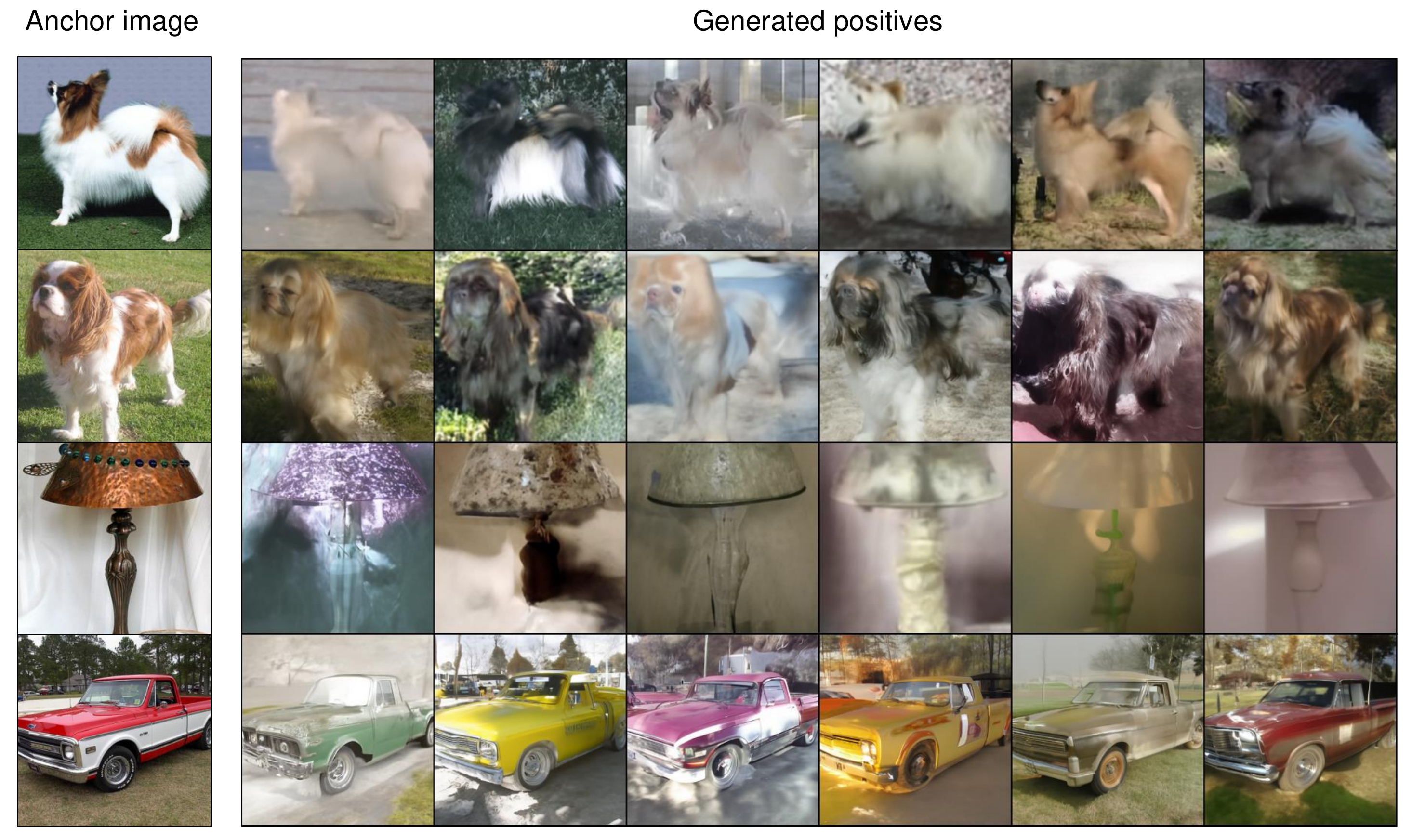}
	\caption{Visualization of generated positives using CLSP on ImageNet100 dataset.}
 \label{fig:generated_positives_imagenet100}
\end{figure}

\begin{table}[t]
\centering
\setlength\tabcolsep{8.0pt}
\caption{Linear evaluation results of different models trained with varied batch sizes on CIFAR10.}
\begin{tabular}{lccccc}
\toprule
Batch size  & 64 & 128 & 256 & 512 & 1024 \\ \hline
SimCLR & 87.61 & 90.34 & 91.01 & 91.32 & 91.47 \\
MoCoV2 & 91.89 & 92.93 & 92.90 & 92.94 & 92.80 \\
CLSP-SimCLR  & 91.18 & 93.36 & 94.04 & 94.26 & 94.37  \\
CLSP-MoCoV2 & 92.55 & 93.43 & 94.35 & 94.41 & 94.10  \\
\bottomrule
\end{tabular}
\label{tab:batch_size}
\end{table}

To evaluate the influence of batch size on the proposed CLSP, we compared the linear evaluation results of SimCLR, MoCoV2, CLSP-SimCLR, and CLSP-MoCoV2 on CIFAR10 under different batch sizes. The results are shown in Table \ref{tab:batch_size}. We can see that even though small batch size degrades the linear evaluation results, our CLSP-SimCLR and CLSP-MoCoV2 consistently outperform the baseline SimCLR and MoCoV2 methods under different batch sizes.

\end{document}